\documentclass[letterpaper, 10 pt, conference]{ieeeconf}  

\IEEEoverridecommandlockouts                              

\usepackage{graphicx}
\usepackage{url}
\usepackage{hyperref}

\title{Skill Sequence Planning for Collaborative Multi-Robot Construction}

\author{
Xi Wang$^{1}$, Bo Fu$^{2}$, Carol C. Menassa$^{3}$, Vineet R. Kamat$^{3}$ and Min Deng$^{4*}$%
\thanks{$^{*}$Corresponding author: Min Deng}
\thanks{$^{1}$Xi Wang is with the Department of Construction Science,
Texas A\&M University, College Station, TX 77843, USA.
{\tt\small xiwang@tamu.edu}}%
\thanks{$^{2}$Bo Fu is with Amazon Robotics, North Reading, MA 01864, USA.
{\tt\small bofu@amazon.com}}%
\thanks{$^{3}$Carol C. Menassa and Vineet R. Kamat are with the Department of Civil and
Environmental Engineering, University of Michigan, MI 48109, USA.
{\tt\small \{menassa, vkamat\}@umich.edu}}%
\thanks{$^{4}$Min Deng is with the Department of Civil and
Environmental Engineering, University of Tennessee, Knoxville,
TN 37996, USA.
{\tt\small mindeng@utk.edu}}%
\thanks{Accepted at the 5th Workshop on Future of Construction: Collaborative Robots for Fabrication, Manufacturing, and Inspection, IEEE/RSJ International Conference on Intelligent Robots and Systems 2026 (IROS 2026).}
}

\begin{document}

\maketitle
\thispagestyle{empty}
\pagestyle{empty}

\begin{abstract}

Robots have significant potential to automate construction processes. However, their industry adoption remains limited, partly because of the programming effort required to adapt robots to diverse tasks. This paper presents a skill sequence planning method that enables a heterogeneous team of multi-functional robots to collaboratively perform construction assembly work using reusable, preprogrammed skills such as grasping, drilling, and fastening. A central controller transforms the digital representation of the building into a construction relationship graph that represents construction entities, their states, and their parent-child relationships. Based on this representation, the system selects the next construction target, generates a symbolic sequence of skills for capable members of the robot team, and produces collision-free geometric motion plans for skill execution. The symbolic planning problem is dynamically regenerated as the construction state changes. An interactive digital twin presents the planned skill sequence and robot states to human co-workers for review and approval before execution. The method is evaluated through a construction assembly case study. By reducing the need to program robots separately for each task variation, the proposed approach supports more flexible deployment of collaborative robot teams in construction.

\end{abstract}

\section{INTRODUCTION}

Construction robots must often perform long-horizon tasks comprising multiple interdependent steps, such as reaching, grasping, fastening, and changing tools \cite{huang2023imitate,wang2023automatic}. Although these individual skills can be preprogrammed and reused, their sequence and parameters vary with the construction target, available materials, jobsite state, and required operation. The planning challenge becomes more complex for a heterogeneous robot team because the system must determine not only which skill should be performed next, but also which robot possesses the capability to perform it \cite{deng2026integrating,chen2026perception}. Manually programming a complete skill sequence for every construction task and operating condition therefore limits the flexible deployment of multi-functional robot teams.

Existing research has investigated several approaches to sequencing robot skills over long horizons. Learning from demonstration allows robots to acquire task knowledge from expert examples instead of explicit programming \cite{ravichandar2020recent}. In construction, Huang et al. combined virtual demonstrations with reinforcement learning for long-horizon window installation \cite{huang2023imitate}, while Wang et al. represented construction work using reusable skill primitives and learned their sequencing through worker demonstrations \cite{wang2023automatic}. However, demonstration-based methods depend on representative demonstrations and may require additional guidance for unseen tasks or states. Large Language Model (LLM)-based methods instead compose predefined skills from high-level instructions. For example, ProgPrompt constrains outputs using available actions and objects \cite{singh2023progprompt}, but such plans still require grounding and verification against execution constraints. Hybrid methods such as LLM+P translate natural-language planning problems into Planning Domain Definition Language (PDDL) representations for classical planning \cite{liu2023llmp}. However, their reliability still depends on the accuracy of the LLM-generated planning representation. Classical symbolic planning based on structured state information avoids this translation uncertainty by explicitly representing action preconditions and effects, while task and motion planning further connects discrete decisions with geometric and kinematic feasibility \cite{garrett2021tamp}.

In construction, skill sequencing must be grounded in project-specific information and continuously updated as physical operations change the jobsite state. This study focuses on dynamically generating valid skill sequences based on the current construction state and the intended installation state, assigning each action to a capable member of a heterogeneous robot team, and connecting the resulting symbolic sequence to collision-free motion execution. The planning representation is also updated as components are installed, materials become accessible, or the actual jobsite state deviates from the expected state.

This paper presents a skill-sequence planning method for collaborative multi-robot construction. A central controller transforms a digital representation of the building into a construction relationship graph that represents construction entities, their states, and their relationships. For each target component, the controller uses the graph to generate a PDDL problem \cite{edelkamp2004pddl22} and combines it with a reusable PDDL domain model that defines robot types, state predicates, and executable skills. A symbolic planner then generates an ordered sequence of parameterized skills in which each action is assigned to a robot with the required capabilities. Motion-related skills, such as robotic-arm reaching, are subsequently converted into collision-free trajectories, while an interactive digital twin allows a human co-worker to review and approve the planned sequence before execution. The proposed method is evaluated through a simulation-based wooden-frame assembly involving two robots with complementary manipulation and fastening capabilities.

\section{METHODOLOGY}

\subsection{System Framework}

\begin{figure}[!b]
  \centering
  \includegraphics[width=\columnwidth]{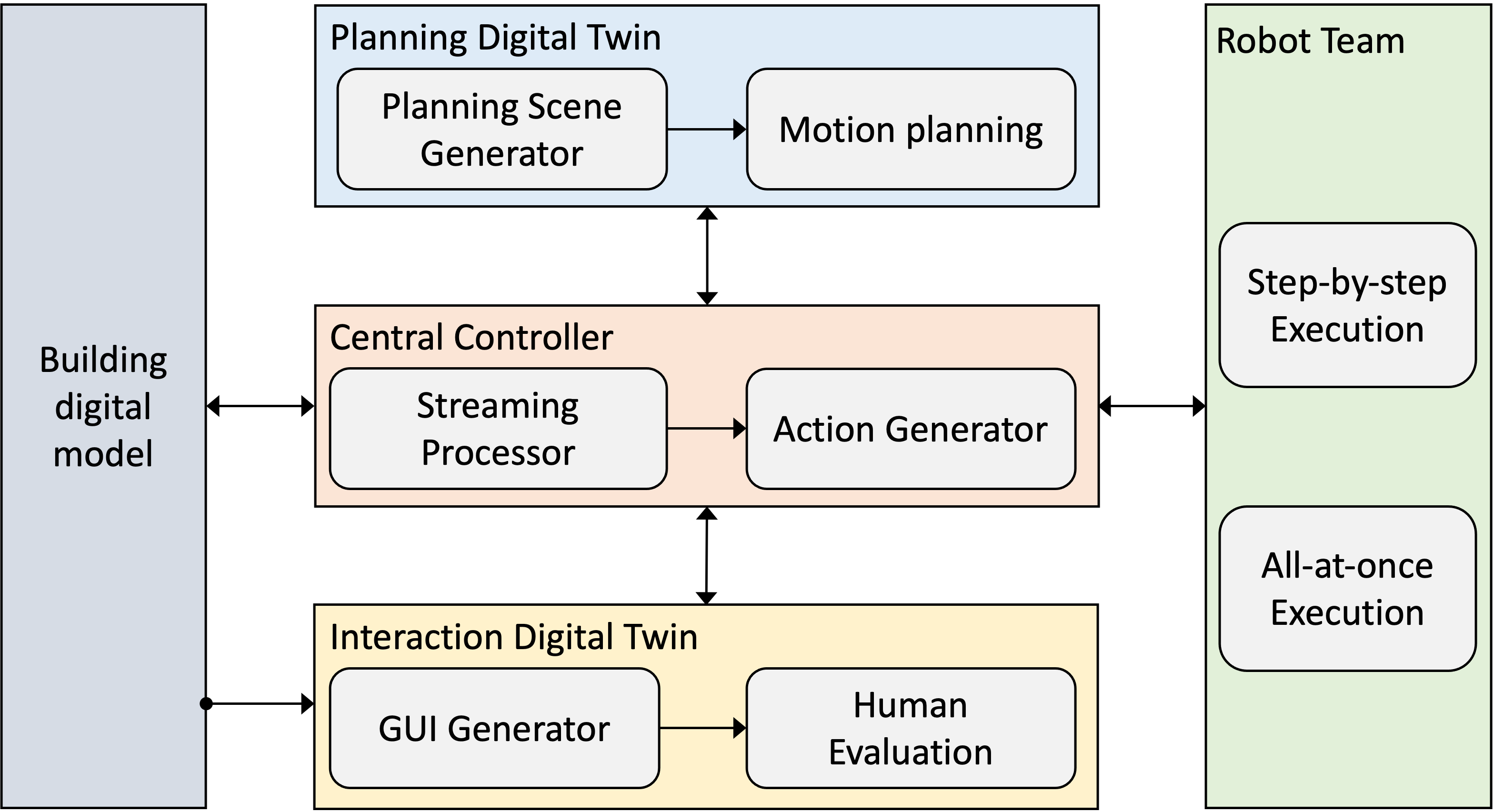}
  \caption{System framework.}
  \label{fig:framework}
\end{figure}

The proposed system enables a heterogeneous team of multi-functional robots to collaboratively perform construction work through five connected modules: an evolving digital building model, a central controller, a planning digital twin, an interaction digital twin, and the robot team (Figure~\ref{fig:framework}). The overall digital twin and communication infrastructure builds on our prior framework \cite{wang2024enabling}. The present work extends this infrastructure with three connected planning levels. First, construction target selection determines the next workpiece to be installed according to the construction sequence stored in a digital model of the building (e.g., BIM). Second, symbolic skill sequence planning determines the ordered robot skills required to complete the selected installation and identifies robot types capable of performing them. Third, geometric motion planning converts each motion-related skill into a collision-free trajectory for the responsible robot.

The digital representation stores the as-designed and as-built information required for robot planning. Its components are organized into Target, Material, As-Built, As-Designed, and Virtual Collision layers. In addition to geometry, the digital model provides attributes such as component type, construction order, world-frame pose, required process, operation method, and construction parent. Target components describe the intended construction state, while Material and As-Built components describe objects currently present on the jobsite. The digital representation is updated as construction progresses.

The central controller retrieves and processes this information, generates skill sequences, and communicates with the other modules through the Robot Operating System (ROS). The planning digital twin uses MoveIt to generate collision-free robot motions from current joint states, target end-effector poses, and workspace geometry. The interaction digital twin, developed in Unity, presents the generated skill sequence to a human co-worker in natural language and visualizes the robot and construction states. The worker can approve the complete sequence for execution or require confirmation before individual actions. Together, these modules establish a human-supervised workflow from construction information to multi-robot execution.

\subsection{Construction State Representation}

The central controller transforms digital building into a representation used to generate a symbolic planning problem. Component geometry and attributes are extracted from the digital model and transmitted to ROS \cite{wang2024enabling}. For each component, the controller creates an object containing its name, geometry, layer, world-frame pose, state, and construction relationships. Components on the As-Built, As-Designed, and Virtual Collision layers are represented by the base \textit{Node} class. The specialized \textit{TargetNode} and \textit{MaterialNode} classes inherit from \textit{Node} and store layer-specific information. A \textit{TargetNode} records the required material type, operations, and target poses, while a \textit{MaterialNode} identifies the type and current location of an available material.

The controller organizes target workpieces and their associated operations into queues using the construction-order attributes provided by the digital model. At each planning cycle, only the target workpiece at the front of the queue and the operations associated with that target are considered. Focusing on one target at a time limits the number of objects, state variables, and actions included in the symbolic planning problem. The remaining components are retained as contextual information for state representation and geometric motion planning. Once the selected workpiece is installed, it is removed from the active queue and the next target becomes available for planning.

\begin{figure}[!b]
  \centering
  \includegraphics[width=\columnwidth]{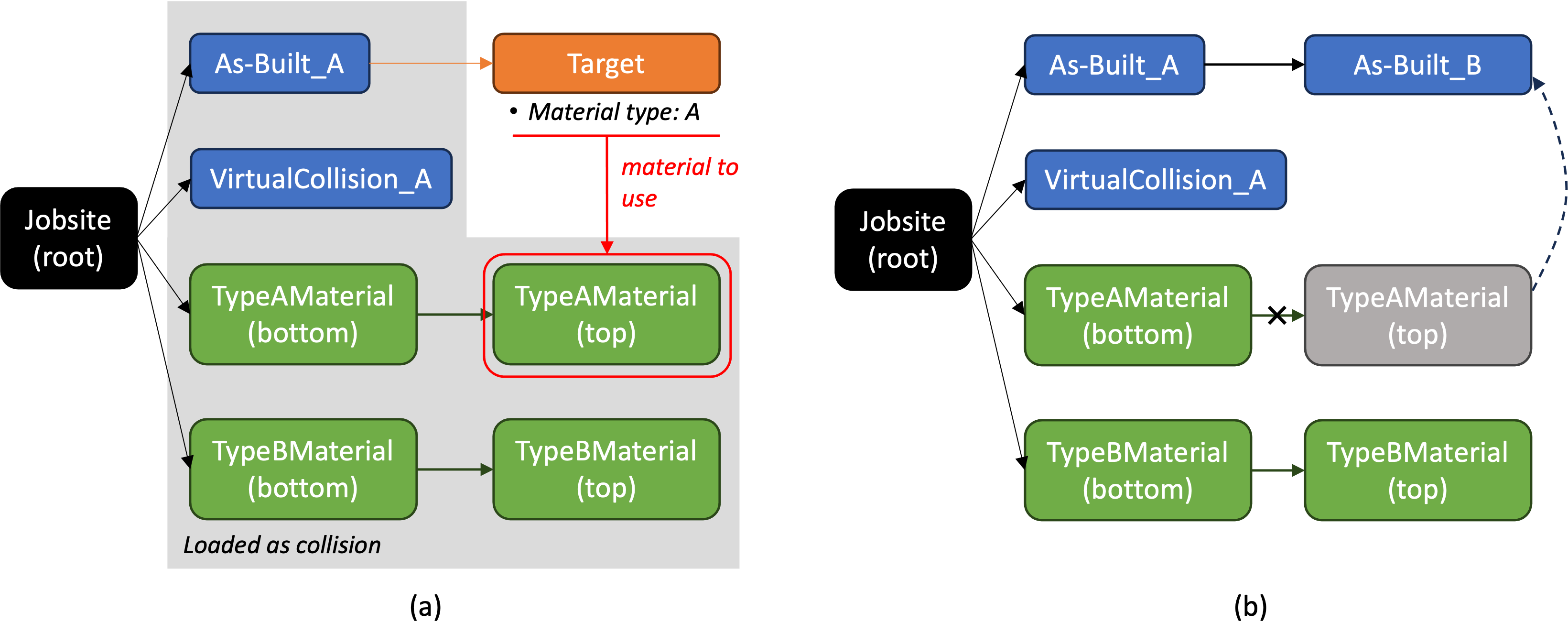}
  \caption{Construction relationship graph (a) graph structure (b) graph update}
  \label{fig:graph}
\end{figure}

To represent relationships among the objects relevant to construction, the controller creates a construction relationship graph, expressed as a directed acyclic graph (Figure~\ref{fig:graph}). Each graph node represents a construction entity and stores its geometry, pose, type, layer, and current state as attributes. Each directed edge represents a construction parent-child relationship. A Jobsite node serves as the default root for components without a specified construction parent.

The graph also represents the availability of materials stored in stacks. When materials of the same type are stacked, the piece below is treated as the construction parent of the piece directly above it. A leaf node therefore represents the currently accessible piece at the top of a stack. For the active target, the controller retrieves the required material type from its attributes and selects the leaf node with the corresponding type. This avoids including inaccessible pieces as candidate materials in the planning problem. If multiple leaf nodes of a corresponding type exist, the one with the shortest Euclidean distance to the responsible robot is chosen.

The construction relationship graph is updated as robot actions change the jobsite state. After a material has been installed, its node changes from a Material component to an As-Built component. Its name and layer are updated, and its parent is changed to the construction parent specified for the corresponding target, or to the Jobsite root if no parent is specified. Removing the installed material also exposes the next piece in its original stack as a new leaf node. The updated graph therefore provides the central controller with the current objects, attributes, and construction relationships required to generate the next symbolic planning problem.

\subsection{Symbolic Skill Sequence Planning}

The central controller formulates skill sequence planning in the PDDL, which separates a reusable domain file from a task-specific problem file \cite{edelkamp2004pddl22}. The controller processes digital building data into the construction relationship graph, uses the graph to generate the PDDL problem, combines the problem with the reusable PDDL domain during planning, and converts the resulting action sequence into both robot commands and natural language descriptions. The domain is manually defined before construction and remains unchanged across targets. The problem file describes the current target and jobsite state, which is is generated automatically for each target and regenerated when the relevant state changes.


\textbf{The domain file }defines \textit{types}, \textit{predicates}, and \textit{actions}. Its type hierarchy organizes planning entities as \textit{Robot}, \textit{Thing}, and \textit{Place}. \textit{Robot} subtypes distinguish elements with different capabilities, \textit{Thing} represents the material or workpiece being manipulated, and \textit{Place} represents symbolic destinations such as the material location, target location, preparation station, and gripper station. Typed predicates describe relevant properties and relationships among these entities, while parameterized actions specify their required parameters, preconditions, and expected effects. Robot subtypes are incorporated into action parameters and preconditions so that each action can be assigned only to a capable team member. Each symbolic action maps to an executable ROS skill, such as reaching, grasping, releasing, fastening, or changing a gripper. Since places are represented symbolically, job-specific coordinates remain outside the domain model. Therefore, their corresponding 6DOF poses are retrieved from the construction relationship graph only when a skill is prepared for motion planning.

\textbf{The problem file} is automatically generated during runtime by grounding the reusable PDDL domain with the current construction representation graph. For each planning cycle, the controller first retrieves the active target from the construction queue and selects an accessible material whose type matches the target requirement. It then generates the \textit{objects} section by instantiating the relevant robots, workpiece, and symbolic places according to the domain types. Robots and recurring places, such as preparation and gripper stations, are treated as persistent objects, whereas the target-specific workpiece and material are retrieved from the graph. After the objects are instantiated, the controller substitutes type-compatible instances into the parameterized predicates defined in the domain to generate candidate grounded state variables. Each variable is evaluated against the graph’s node attributes, construction parent-child relationships, and current robot states. Variables that evaluated as true are written to the \textit{initial state}. The same predicates are then evaluated for the intended post-installation condition represented by the active target, and the resulting true variables form the \textit{goal state}. These three sections are serialized into a PDDL problem file. When the active target or observed construction state changes, the controller regenerates the file so that subsequent skill sequence planning reflects the updated jobsite condition.

The domain and problem files are submitted to Fast Downward, which uses heuristic search to identify a valid sequence of parameterized actions \cite{helmert2006fast}. The plan specifies the skills, their object and place parameters, and the capable robot associated with each action. Because this output is symbolic text, the controller applies two processors. One converts the actions into corresponding ROS functions and control parameters. The other translates them into natural-language descriptions for review in the interaction digital twin.

For a motion-related skill, the planning digital twin resolves its symbolic place into a current target pose and uses MoveIt to generate a collision-free trajectory. Geometry from the Material, As-Built, and Virtual Collision layers, together with the other robots, is considered for colision avoidance. The interaction digital twin presents the sequence for human approval. If the worker selects a different feasible action, that action is executed, the observed construction state is updated, and a new PDDL problem and skill sequence are generated. If geometric motion planning fails to identify a feasible trajectory to the target, the failure is escalated to the human co-worker. After the target is completed, the digital model and relationship graph are updated and the workflow advances to the next target.

\section{CASE STUDY}

A simulation-based case study was conducted as a proof-of-concept implementation to verify that the proposed method could translate a building digital model into an executable, human-supervised skill sequence for a heterogeneous robot team. The study focused on the end-to-end integration of target selection, symbolic skill sequence planning, robot assignment, geometric motion planning, and human approval. Two KUKA KR 120 manipulators with different end-effectors were used in the case study. One robot is responsible for workpiece manipulation, including picking, transporting, positioning, and releasing each component. The other robot is responsible for connection or finishing operations enabled by exchangable tools, including nailing, screwing, and painting.

\begin{figure}[!b]
  \centering
  \includegraphics[width=\columnwidth]{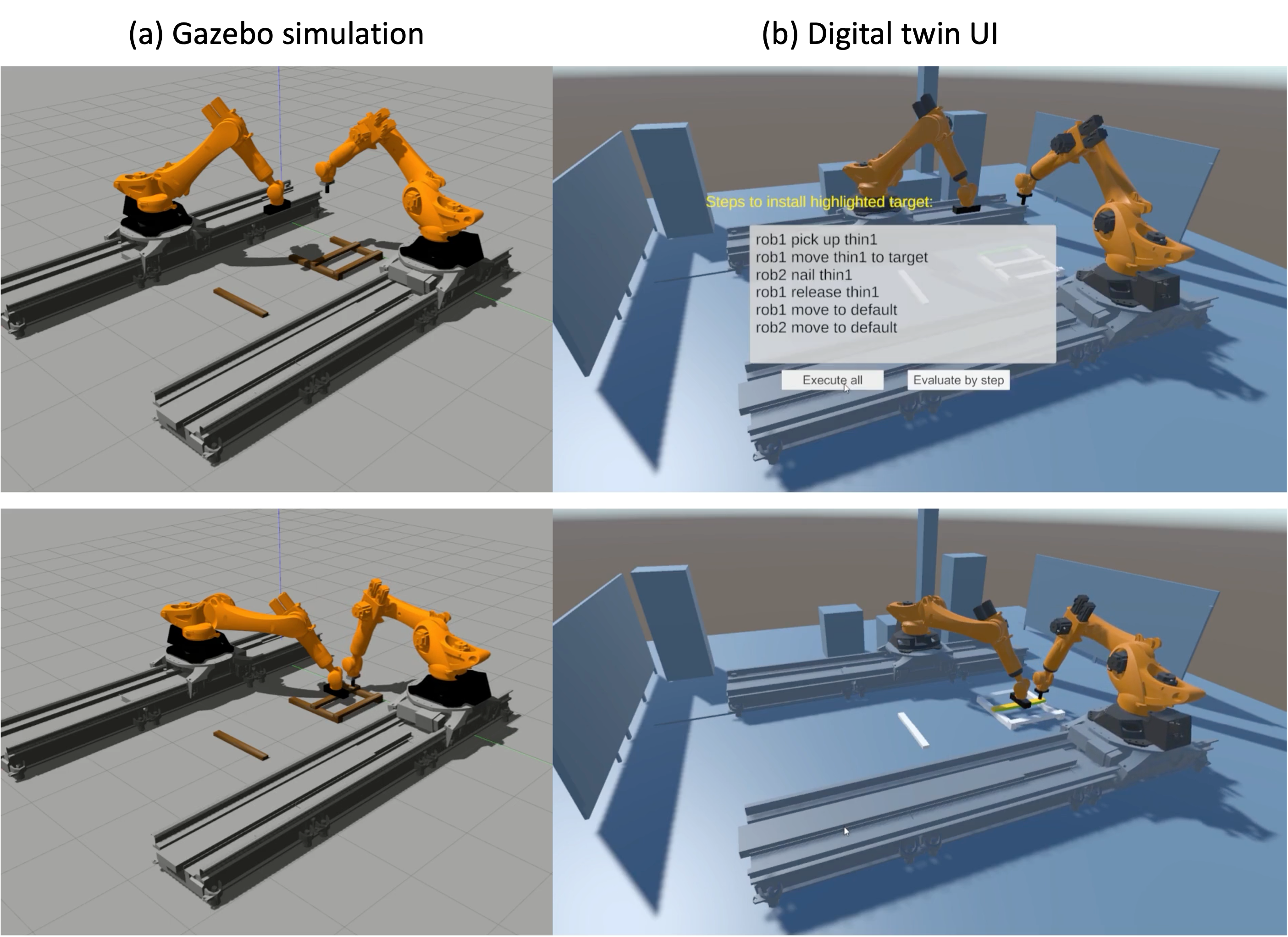}
  \caption{Snapshots of the wooden frame installation case study.}
  \label{fig:planning}
\end{figure}

The case study demonstrated the assembly of a wooden frame comprising two thicker supporting studs and three thinner transverse studs. The supporting studs were positioned first. Each transverse stud was subsequently placed across the supports and secured using four nails, with two nails at each end. The construction environment was simulated in Gazebo, as shown in Figure~\ref{fig:planning}(a). The interaction digital twin was developed in Unity and communicated with the simulation through ROSbridge to exchange robot, construction, and planning information. The case study uses ground-truth simulation states for planning. Although the underlying digital twin framework supports sensor-based state updates \cite{wang2024enabling}, recovery from sensor-detected execution deviations is not evaluated here. The building digital model was maintained in Rhino, with semantic information attached to the geometric components as attribute data. As illustrated in Figure~\ref{fig:planning}(b), the interface visualized the evolving construction state and presented the planned skills in natural language for human review.

For each target component, the controller retrieved its construction order, target pose, required material type, construction parent, and connection requirements from the building digital model. It then generated a PDDL problem, selected a capable robot for each skill, and produced collision-free motions for execution. The resulting sequence could be approved and executed as a complete plan or evaluated one step at a time through the Unity interface. In both modes, the simulated robots completed the planned manipulation and nailing sequence while their states were reflected in the interaction digital twin. The two execution modes are demonstrated in the supplementary videos: \href{https://youtu.be/4UcvW9Bcb3w}{complete plan execution} and \href{https://youtu.be/xKZCsYBkN6U}{execution with step-by-step approval}. These results provide qualitative evidence that the proposed representation and planning method can coordinate multiple robots and support human-supervised execution of a multi-stage construction assembly task in simulation.

\section{CONCLUSIONS}

This paper presented a skill sequence planning method that connects evolving construction information to human-supervised execution by a heterogeneous robot team. A construction relationship graph represents construction entities, their states, material accessibility, and parent-child relationships, enabling the central controller to automatically generate and update target-specific PDDL problems. Combined with a reusable domain model, the method generates ordered skill sequences, assigns actions to robots with the required capabilities, and converts motion-related skills into collision-free trajectories. The wooden-frame case study demonstrated the end-to-end integration of target selection, symbolic planning, robot assignment, geometric motion planning, and human approval. In simulation, two robots with complementary capabilities completed the planned manipulation and nailing operations under both complete-plan and step-by-step approval modes. Limitations of the current method include planning skills for one target at a time, assuming global assembly dependencies are provided by the building model, and supporting sequential rather than concurrent multi-robot execution. Future work will establish quantitative evidence of the method's feasibility, validate the method with physical robots, and evaluate its scalability across larger robot teams and more diverse construction processes. 

\addtolength{\textheight}{-12cm}   




\bibliographystyle{IEEEtran}

\bibliography{ref}

\end{document}